\documentclass[10pt,twocolumn,letterpaper]{article}

\usepackage{cvpr}              

\usepackage[dvipsnames]{xcolor}

\usepackage{subcaption}
\usepackage{setspace}
\usepackage{color}
\usepackage{colortbl}
\usepackage{xcolor}
\usepackage{amsmath}
\usepackage{multirow}
\newlength\savewidth\newcommand\shline{\noalign{\global\savewidth\arrayrulewidth
  \global\arrayrulewidth 1pt}\hline\noalign{\global\arrayrulewidth\savewidth}}

\definecolor{mygray1}{gray}{.9}
\definecolor{mygray2}{gray}{.7}
\definecolor{cvprblue}{rgb}{0.21,0.49,0.74}
\usepackage[accsupp]{axessibility} 
\usepackage[pagebackref,breaklinks,colorlinks,citecolor=cvprblue]{hyperref}

\newcommand{\app}{\raise.17ex\hbox{$\scriptstyle\sim$}}
\def\eg{\emph{e.g}\onedot} 
\def\ie{\emph{i.e}\onedot}

\def\etal{\emph{et al}\onedot}
\makeatother

\def\paperID{13521} 
\def\confName{CVPR}
\def\confYear{2024}

\makeatletter\renewcommand\paragraph{\@startsection{paragraph}{4}{\z@}
	{.5em \@plus1ex \@minus.2ex}{-.5em}{\normalfont\normalsize\bfseries}}\makeatother
\title{Revisiting Adversarial Training at Scale}

\author{%
Zeyu Wang\textsuperscript{*} \quad
Xianhang Li\textsuperscript{*} \quad
Hongru Zhu \quad 
Cihang Xie \vspace{.3em}
\\
\small $^{*}$equal contribution \vspace{.5em} \\
UC Santa Cruz  \\
}

\begin{document}
\maketitle
\begin{abstract}
The machine learning community has witnessed a drastic change in the training pipeline, pivoted by those ``foundation models'' with unprecedented scales. However, the field of adversarial training is lagging behind, predominantly centered around small model sizes like ResNet-50, and tiny and low-resolution datasets like CIFAR-10. To bridge this transformation gap, this paper provides a modern re-examination with adversarial training, investigating its potential benefits when applied at scale. Additionally, we introduce an efficient and effective training strategy to enable adversarial training with giant models and web-scale data at an affordable computing cost. We denote this newly introduced framework as AdvXL.

Empirical results demonstrate that AdvXL establishes new state-of-the-art robust accuracy records under AutoAttack on ImageNet-1K. For example, by training on DataComp-1B dataset, our AdvXL empowers a vanilla ViT-g model to substantially surpass the previous records of $l_{\infty}$-, $l_{2}$-, and $l_{1}$-robust accuracy by margins of \textbf{11.4\%}, \textbf{14.2\%} and \textbf{12.9\%}, respectively. This achievement posits AdvXL as a pioneering approach, charting a new trajectory for the efficient training of robust visual representations at significantly larger scales. Our code is available at \url{https://github.com/UCSC-VLAA/AdvXL}.
\end{abstract}    
\section{Introduction}
\label{sec:intro}

The landscape of machine learning, particularly deep learning, has witnessed a transformative shift with the advent of large-scale models and datasets. This paradigmatic shift, exemplified by the inception of ``foundation models'' such as Large Language Models (LLMs) \cite{devlin2018bert,brown2020language,touvron2023llama,touvron2023llamav2,gpt4}, has redefined the boundaries of what is achievable in various domains of artificial intelligence. 
Excitingly, parallel developments have also been observed in computer vision --- recent advancements in scaling datasets and model sizes have mirrored the feasibility of ``LLM-like'' scaling for building exceptionally strong visual recognition models~\cite{eva,zhai2022scaling,dehghani2023scaling}.

\begin{figure}[tbh!]
  \centering
    \begin{subfigure}[b]{0.7\linewidth}
        \centering
          \includegraphics[width=1.0\textwidth]{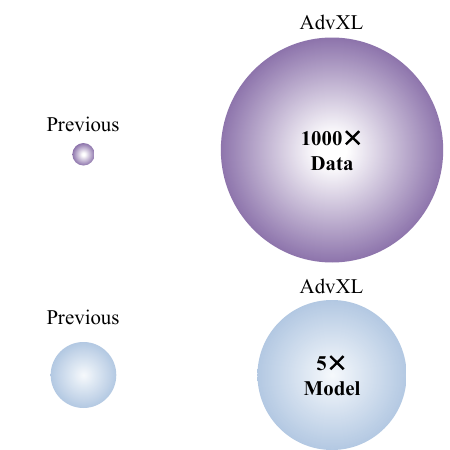}
        \caption{Scale comparison.}
        \label{fig:teaser scale}
    \end{subfigure}
    \begin{subfigure}[b]{0.8\linewidth}
        \centering
        \includegraphics[width=\linewidth]{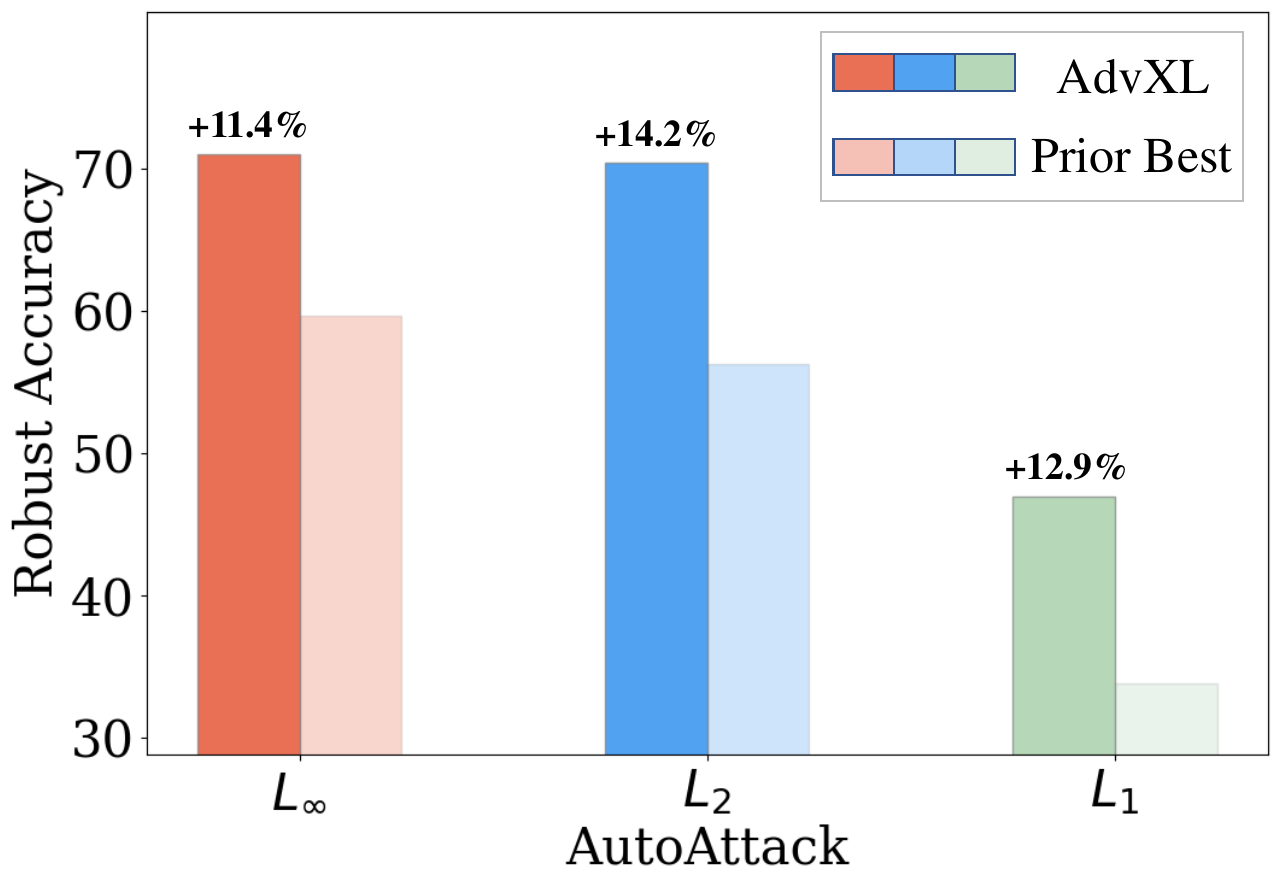}
        \caption{Performance comparison.}
        \label{fig:performance highlight}
    \end{subfigure}
    \vspace{-1em}
    \caption{Our AdvXL increases significantly in terms of both model size and data scale, which brings a substantial boost over prior best results of $l_{\infty}$, $l_{2}$, and $l_{1}$ robustness on ImageNet-1K, even though our model is only trained to be $l_{\infty}$-robust.}
    \label{fig:teaser}
    \vspace{-1.5em}
\end{figure}

However, amidst this evolution, adversarial training~\cite{madry2018towards,goodfellow2014explaining} --- a pivotal strategy aimed at securing model robustness against adversarial attacks --- has faced significant scalability challenges in this foundation model era. Adversarial training, typically employed in small models such as ResNet-50 \cite{he2016deep} trained on small datasets like CIFAR-10 \cite{Krizhevsky09learningmultiple}, involves repeatedly generating adversarial examples through on-the-fly attacks during the training process. This iterative and intensive procedure demands substantial computational resources, thus making it challenging to scale up.

Contrasting with these challenges, recent endeavors in adversarial training have indeed shown intriguing glimpses of promise from data scaling by incorporating 50 million additional images to sustain state-of-the-art robustness records on CIFAR-10~\cite{wang2023better}. Additionally, other adversarial training works~\cite{singh2023revisiting,liu2023comprehensive} attain impressive performance with model scaling using larger models like Swin-L~\cite{liu2021swin} and ConvNeXt-L~\cite{liu2022convnet} on ImageNet-1K. These observations, coupled with the burgeoning success of foundation models, instigates a critical question: \emph{can the principles of model and data scaling, already proven effective in vanilla training, be transferable to adversarial training?} Moreover, how effectively does such scaling translate to robustness improvement in adversarial training?

In response to these questions, we re-examine adversarial training at a previously uncharted foundation-model scale. 
In terms of model scaling, we increased the model parameters from the previously largest 200M size to $\textbf{1B}$; 
for data scaling, we adversarially train models on various datasets spanning from the medium-size ImageNet-1K with around 1M images to the web-scale dataset comprising more than  \textbf{1B} images. Additionally, to make the scaling of adversarial training computationally affordable, we introduce an efficient approach with a straightforward two-stage training schedule, \ie, first lightweight pre-training, then intensive fine-tuning. We name this efficient and scalable adversarial training framework as AdvXL.

Collectively, extensive experiments showcase that these scaling endeavors successfully result in substantial improvements over the previous state-of-the-art methods on adversarial robustness. 
For example, by training a one-billion parameter model on a one-billion image dataset, we establish a new state-of-the-art record for $l_{\infty}$-robust accuracy of 71.0\% under AutoAttack on ImageNet-1K, marking a substantial enhancement in model robustness. Notably, AdvXL demonstrates exceptional generalizability when tested against unseen attacks, improving upon the previous best $l_2$- and $l_1$-robust accuracy of models trained to be $l_{\infty}$-robust by margins of $\app$14\% and $\app$13\%, respectively. These results underscore the pivotal role of (significantly) scaled adversarial training in enhancing model robustness against diverse adversarial threats.

\section{Related Work}
\label{sec:related_work}

\subsection{Adversarial Training}
Adversarial training has emerged as a pivotal defense mechanism against adversarial attacks in machine learning. Initially introduced by Goodfellow~\etal~\cite{goodfellow2014explaining}, this methodology involves training models on crafted adversarial examples designed to provoke model misclassification. Subsequent studies have extended this foundation, examining facets such as the impact of batch size, learning rate, data augmentation, and training duration on model robustness, predominantly on smaller datasets like CIFAR-10 \cite{gowal2020uncovering, pang2021bag,li2022bag,mo2022adversarial, hendrycks2019using, chen2020adversarial}. Other research efforts have explored deeper nuances of adversarial training recipes tailored for ImageNet-1K~\cite{singh2023revisiting,bai2021transformers,debenedetti2023light,rebuffi2023revisiting,Xie_2019_CVPR,xie2020intriguing,xie2020smooth}. Recent works also investigate the robustness of novel network designs like Vision Transformer (ViT) \cite{fu2022patch,singh2023revisiting,gu2022vision,croce2021robustbench}. In particular, Singh \etal~\cite{singh2023revisiting} achieve the best generalized robustness by enhancing ViT and ConvNeXT with Convolutional Stem.

Despite its effectiveness, adversarial training is notoriously resource-intensive, limiting its scalability. To address this challenge, researchers have pursued more resource-efficient adversarial training methodologies. Examples include Free Adversarial Training~\cite{shafahi2019adversarial} and Fast Adversarial Training~\cite{wong2020fast}, both aimed at reducing training costs while preserving model robustness. 
However, these approaches have predominantly focused on smaller networks and datasets, leaving a noticeable gap concerning large-scale models. In this work, we aim to significantly expand the horizons of scaling adversarial training to unprecedented levels of efficiency and effectiveness.

\subsection{Scaling Vision Foundation Models}
Parallel to large-scale language models, exemplified by innovations like GPT series~\cite{gpt4}, similar efforts have been made for vision models, particularly with the scaling of ViTs~\cite{zhai2022scaling, eva, dehghani2023scaling}.
Liu~\etal~\cite{liu2022swin} effectively trained the SwinV2-G model, housing an astounding 3B parameters, by employing residual-post-norm and scaled cosine attention.
Similarly, Dehghani ~\etal~\cite{dehghani2023scaling} have shown substantial performance enhancements by scaling ViTs to 22B parameters, mirroring the scaling trends witnessed in language models.

Despite the burgeoning scaling efforts in vision foundation models, the exploration of adversarial training has traditionally been limited to small or base model sizes.
Recent scaling effort has led to noteworthy performance improvements, evidenced by the achievements on RobustBench~\cite{croce2021robustbench} with larger models like Swin-L and ConvNeXt-L~\cite{liu2023comprehensive, singh2023revisiting}. 
Diverging from these antecedent initiatives, our work explores adversarial training at an even much larger scale, up to the training of a one-billion-parameter model on one-billion samples, thereby pioneering the frontiers of adversarial training into uncharted territory.
\section{AdvXL}
\label{sec:method}

In this section, we introduce AdvXL, a novel training framework designed for adversarially robust visual representation learning at scale. We first revisit the fundamental concept of adversarial attacks and adversarial training in Sec.~\ref{sec:adv train}. Following this, in Sec.~\ref{sec:two-stage training}, we present a two-stage efficient adversarial training pipeline characterized by a coarse-to-fine, weak-to-strong approach. In Sec.~\ref{sec:cLIP embedding for adv train}, we showcase how to leverage CLIP~\cite{radford2021learning} text encoder as a tool for enabling us to learn with web-crawled images, where a precise label is usually missing but with a corresponding text description, for scaled adversarial training.

\subsection{Adversarial Training}
\label{sec:adv train}
Adversarial examples are uniquely crafted inputs that, despite their visual similarity to authentic samples within specific norm constraints, are engineered to deceive machine learning models into producing inaccurate predictions. These examples play a crucial role in assessing the robustness of a model in scenarios where malicious manipulations may occur.

Adversarial Training is central to fortifying a model against such adversarial inputs. This technique involves a strategic training process designed to enhance the model's robustness to adversarial attacks. The mathematical foundation of AT is encapsulated as an optimization problem:

\begin{equation}
\min _\theta \sum_{\left(x_i, y_i\right) \in \mathcal{D}} \max _{\delta:\|\delta\|_p \leq \epsilon_p} \mathcal{L}\left(f_\theta\left(x_i+\delta\right), y_i\right),
\label{eq:adv train}
\end{equation}
where $\theta$ represents the parameters for a network $f_{\theta}$. The objective is to train the network $f_{\theta}$ such that it maintains consistent predictions under adversarial perturbations $\delta$, \ie, within an $l_{p}$-ball of radius $\epsilon_{p}$ centered around each input sample $x_{i}$. 

Adversarial Training has proven highly effective to safeguard models against adversarial threats~\cite{biggio2013evasion,szegedy2013intriguing,goodfellow2014explaining}. In our approach, we adopt the widely recognized PGD-based Adversarial Training (PGD-AT) method for the inner maximization problem, renowned for its robust performance and computational efficiency. For the outer minimization problem, we typically employ optimization algorithms like Stochastic Gradient Descent or AdamW \cite{loshchilov2018decoupled}, using cross-entropy as the loss function $\mathcal{L}$.

\begin{figure*}[htb]
  \centering
  \includegraphics[width=0.7\linewidth]{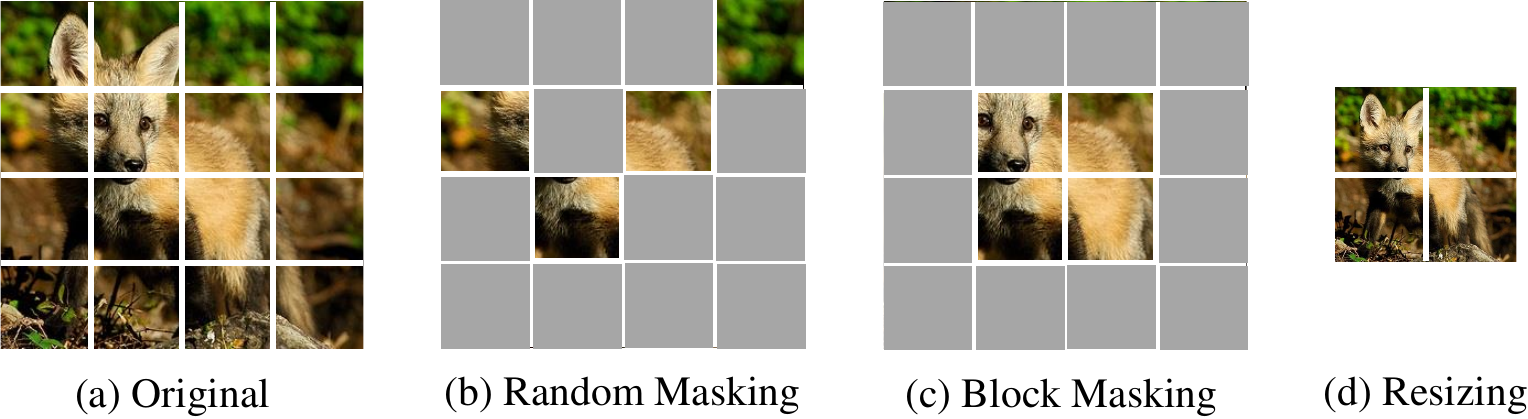}
  \vspace{-.8em}
   \caption{Illustration of different approaches to image token reduction.}
   \label{fig:token_reduction}
   \vspace{-1em}
\end{figure*}

\subsection{Two-stage Training}
\label{sec:two-stage training}

Our adversarial training framework hinges on a two-stage process: a lightweight pre-training stage and an intensive fine-tuning stage. During the pre-training stage, the model is trained with inputs at reduced token length and weaker attacks, spanning a relatively extended duration. Then, during the subsequent fine-tuning stage, the model is trained with inputs at full resolution and stronger attacks, following a comparatively shorter schedule. 
Compared to the vanilla one-stage adversarial training pipeline, this coarse-to-fine (\emph{w.r.t.} input), weak-to-strong (\emph{w.r.t.} adversarial attacker), two-stage training pipeline significantly reduces the overall training cost, rendering it computationally affordable for further scaling up.

\paragraph{Coarse-to-fine training.} We first explore various strategies for image token reduction in the initial pre-training stage. Following \cite{li2023clipav2,li2023inverse}, three distinct approaches are investigated:
\begin{itemize}
    \item \textit{Random Masking}. This method, as described in \cite{he2022masked,li2023scaling}, involves dividing an image into non-overlapping patches (\eg, 16$\times$16), subsequently masking a random proportion of these patches (\eg, 75\%). The model only processes the visible patches, reducing the computational cost by 50\% or 75\%, depending on the masking ratio.
    
    \item \textit{Block Masking}. Inspired by \cite{bao2022beit}, this approach retains tokens from a consecutive large block within the image while discarding others. This method leverages the common placement of objects in the central regions of images, potentially preserving semantic meaningful tokens while significantly reducing the computational cost from lengthy inputs.
    
    \item  \textit{Resizing}. Image resizing is another method for reducing the image token length. Compared to masking, resizing retains more image information, especially high-level semantics. For instance, resizing an image to $112\times112$ is computationally akin to applying a 75\% masking ratio to an image resized to $224\times224$. In our approach, we choose anti-aliasing bilinear interpolation to better preserve the image quality.
\end{itemize}
A visual comparison illustrating these image token reduction strategies is presented in Fig.~\ref{fig:token_reduction}. These strategies are evaluated to discern their efficacy in achieving training acceleration while retaining critical image semantics.

\paragraph{Weak-to-strong training.} Another critical factor for accelerating adversarial training involves managing the number of gradient steps used to craft adversarial samples. Generally speaking, increasing the number of gradient steps results in stronger attacks and enhances adversarial robustness but inevitably inflates computational costs. It has been reported that forming a robust network with adversarial training can take significantly longer, ranging from 3 to 30 times more than building a non-robust equivalent~\cite{shafahi2019adversarial}. As a result, previous studies~\cite{poursaeed2018generative,xiao2018generating,shafahi2019adversarial} have proposed strategies like recycling gradient information or employing a small generator network to mitigate the significant computational burden in adversarial training. 

Our exploration reveals that applying a small number of PGD steps (\eg, PGD-1) during the pre-training stage and subsequently increasing these steps during the fine-tuning phase (\eg, PGD-3) sufficiently secure strong robustness, \ie, this method proves effective compared to initiating training with strong attacks. Importantly, this approach contributes a notable additional speedup, enhancing the efficiency gained from the coarse-to-fine training pipeline (\eg, up to $2\times$),
as solving the inner optimization of adversarial training often requires optimization with multiple iterations and is extremely time-consuming.

\paragraph{Fine-tuning.} Echoing findings from prior research \cite{li2023scaling,li2023inverse}, we find that further adversarially training our model with full-resolution inputs and stronger attacks for a short schedule yields considerable improvement and delivers a more favorable accuracy-to-time trade-off. Compared to the pre-training stage, the fine-tuning phase is notably shorter, often reduced by one or two orders of magnitude. Therefore, even though each sample may entail a notably higher number of image tokens (\eg, $4\times$ by switching back to full image resolution) and require more gradient steps (\eg, $2\times$ by switching back to the strong PGD-3 attacker) in this fine-tuning phase, the overall computation does not increase significantly.

\begin{figure*}[htb]
  \centering
  \includegraphics[width=0.78\linewidth]{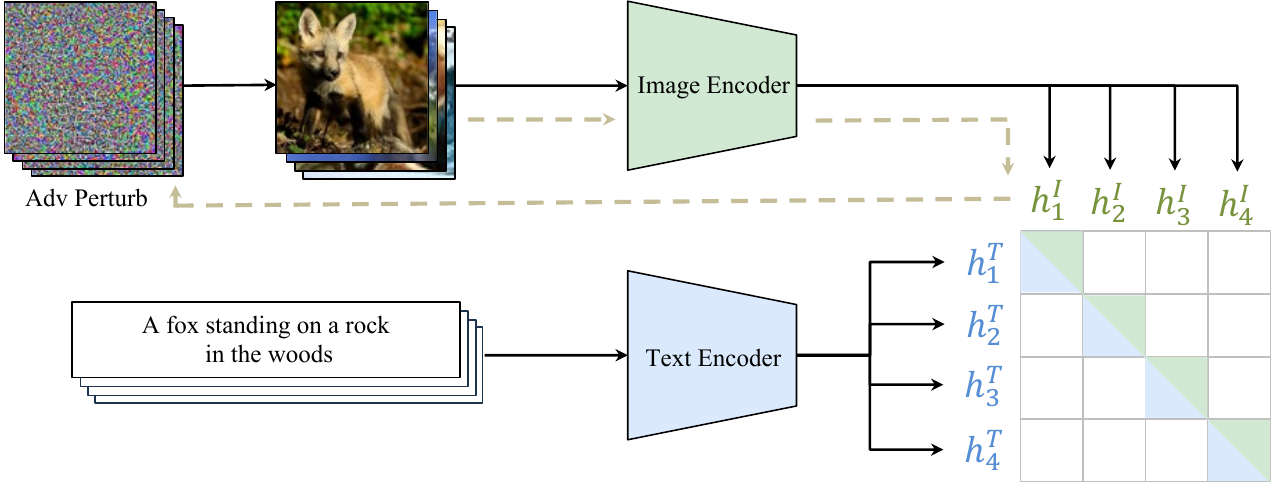}
  \vspace{-1.2em}
   \caption{Illustration of leveraging CLIP embedding in adversarial training. The gray line denotes the adversarial example generation flow.}
   \label{fig:adv_clip}
   \vspace{-1em}
\end{figure*}

\subsection{CLIP Embedding for Web-Crawled Images}
\label{sec:cLIP embedding for adv train}

Previous works have leveraged the zero-shot generalization capability of pre-trained CLIP text encoder~\cite{radford2021learning} to aid a range of downstream tasks, including object detection \cite{gu2022openvocabulary,zhong2022regionclip,zhou2022detecting} and segmentation~\cite{li2022languagedriven,rao2022denseclip} in an open-vocabulary setting. 
Similarly, we hereby propose to employ CLIP text encoder to extract classifier weights when training on web-crawled large-scale datasets with open text descriptions, such as LAION-400M~\cite{schuhmann2021laion} and DataComp-1B~\cite{gadre2023datacomp}. Moreover, adversarial training on these gigantic datasets enables the model to transcend pre-defined categories and directly learn intricate class relationships through natural language supervision. 

Specifically, we adopt the contrastive loss from~\cite{tian2020contrastive,radford2021learning}, formulated as:
\begingroup
\makeatletter\def\f@size{8}\check@mathfonts
\def\maketag@@@#1{\hbox{\m@th\large\normalfont#1}}
\begin{equation}
\begin{split}
& \mathcal{L} \left(f^I, f^T, I_{i}, T_{i}  \right) =  \\
& -\frac{1}{2 n} \sum_i\left(\log \frac{\exp \left(h_i^T \cdot h_i^I / \tau\right)}{\sum_j \exp \left(h_i^T \cdot h_j^I / \tau\right)}+\log \frac{\exp \left(h_i^I \cdot h_i^T / \tau\right)}{\sum_j \exp \left(h_i^I \cdot h_j^T / \tau\right)}\right)
\end{split}
\label{eq:contrastive loss}
\end{equation}
\endgroup
where $n$ represents the batch size; $\tau$ is a learnable temperature parameter; $h_i^I=f^I\left(I_i\right) /\left|f^I\left(I_i\right)\right|$ and $h_i^T=f^T\left(T_i\right) /\left|f^T\left(T_i\right)\right|$ denote the normalized projected features of an image-text pair ($I_i$, $T_i$). Note that we opt for CLIPA-trained text encoder~\cite{li2023inverse,li2023clipav2} as the initial $f^T$ weight and keep it frozen during training. In this case, the adversarial training framework can be described as the following optimization problem,
\begin{equation}
\min _{\theta^I} \sum_{\left(I_i, T_i\right) \in \mathcal{D}} \max _{\delta:\|\delta\|_p \leq \epsilon_p} \mathcal{L} \left(f^I, f^T, I_{i}+\delta, T_{i}  \right),
\label{eq:contrastive adv train}
\end{equation}
where $\theta^I$ represents the parameters of the image encoder $f^I$. 
To elucidate this integration further, Fig.~\ref{fig:adv_clip} provides a visual representation illustrating the incorporation of the CLIP encoder in adversarial training.
\section{Experiment}
\label{sec:experiment}
In this section, we first introduce the datasets used for adversarial training, along with the details of the training and evaluation setup in Sec.~\ref{sec:implementation}. In Sec.~\ref{sec:design choices}, we delve into the ablation results, exploring key elements in our two-stage training pipeline. Furthermore, we investigate the performance of adversarial training as the model, data, and schedule scale synergistically in Sec.~\ref{sec:scaling behavior}. Finally, we compare and contrast the efficiency and efficacy of AdvXL against prior arts in Sec.~\ref{sec:sota comparison}.
\subsection{Implementation}
\label{sec:implementation}

\paragraph{Dataset.} We utilize four different datasets as the training set for adversarial training, which are ImageNet-1K and ImageNet-21K~\cite{deng2009imagenet} — two well-curated labeled datasets for supervised training, as well as LAION-400M~\cite{schuhmann2021laion} and DataComp-1B~\cite{gadre2023datacomp} — two weakly labeled datasets with natural language captions crawled from the Internet.

Specifically, ImageNet-1K comprises approximately 1.28M images from 1000 classes, while ImageNet-21K consists of around 13M images from 19k classes. LAION-400M is the first publicly available web-scale dataset consisting of 400M image-text pairs. It is filtered by CLIP and NSFW criterion, but is still relatively non-curated. DataComp-1B is a more recent dataset with about 1.3B samples filtered from a candidate pool of 12.8B image-text pairs from Common Crawl, which has been recorded to yield superior performance for contrastive training. 

To summarize, our choices of training datasets cover a wide range of representative datasets, spanning from $\app$1M to $\app$1B samples, from well-curated labeled data to non-curated web data. This diverse selection enables a comprehensive investigation into the adversarial training concerning data scaling behaviors.

\paragraph{Training.} By default, our training initiates with a pre-training stage utilizing an image size of $112\times112$ and PGD-1 with a step size of 4/255. Subsequently, the model undergoes a fine-tuning stage employing an image size of $224\times224$ and PGD-3 with a step size of 4/255. 
Our primary focus centers on ViT~\cite{dosovitskiy2021an}, renowned for its scalability \cite{dosovitskiy2021an,radford2021learning,he2022masked,li2023scaling} yet relatively underexplored in the realm of adversarial training. Note that the current best ViT model on ImageNet-1K in RobustBench is only ViT-B/16~\cite{croce2021robustbench}, indicating plenty of room for further scaling.

On ImageNet-1K and ImageNet-21K, our recipe closely follows prior works~\cite{he2022masked}, which successfully trains ViTs on ImageNet at scale from scratch. Specifically, we adopt the AdamW optimizer~\cite{loshchilov2018decoupled} with a short-term linear learning rate warmup followed by a cosine learning rate schedule. Our data augmentation strategy integrates RandAug~\cite{cubuk2020advances},
MixUp~\cite{zhang2018mixup} and CutMix~\cite{yun2019cutmix}. Additionally, we incorporate stochastic depth~\cite{huang2016deep} and weight decay for model regularization. On web-scale datasets such as LAION-400M and DataComp-1B, our training recipe aligns with methodologies outlined in~\cite{li2023inverse}.

The specifics of our training schedules are tailored to individual datasets, where the total number of training samples serves as the primary metric, following a paradigm akin to CLIP training~\cite{radford2021learning,li2023scaling,li2023inverse}. For instance, our default pre-training schedule on ImageNet-1K spans a total of 256M samples, which corresponds to 200 epochs of training.

\paragraph{Evaluation.} In our analysis, we primarily use robust accuracy under PGD-20 attack with a step size of 1/255 as the principal metric. When comparing against other state-of-the-art methods, we follow RobustBench~\cite{croce2021robustbench} and use the robust accuracy evaluated on a subset of selected 5000 images of the ImageNet-1K validation set under AutoAttack. AutoAttack is a standardized adversarial robustness benchmark that consists of an ensemble of white- and black-box attacks, including APGD for cross-entropy and targeted DLR loss,
FAB-attack~\cite{croce2020minimally} and the black-box Square Attack~\cite{andriushchenko2020square}. The attack radii are $\epsilon_{\infty}$ = 4/255, $\epsilon_{2}$ = 2, and $\epsilon_{1}$ = 75 for $l_{\infty}$, $l_2$, and $l_1$ attacks, respectively.

\begingroup
\renewcommand{\arraystretch}{1.2} 
\begin{table}[tbp]
\centering
\vspace{.5em}
\begin{subtable}[ht]{0.43\textwidth}
    \resizebox{\linewidth}{!}{
    \begin{tabular}{ccc|cc}
    Approach & Ratio/Size & Compute & Clean & PGD-20 \\
    \shline
    baseline & 224/0\%    &1.0$\times$ &75.5   &54.5    \\ \hline
    Random Masking & 224/50\%    &0.5$\times$  &72.0   &51.9    \\
    Random Masking & 224/75\%    &0.25$\times$ &67.3   &46.5    \\ \hline
    Block Masking & 224/50\%    &0.5$\times$ &72.3   &52.0    \\
    Block Masking & 224/75\%    &0.25$\times$ &70.6   &49.3    \\ \hline
    Resizing & 160/0\%    &0.5$\times$ &74.7   &53.9    \\
    \rowcolor{mygray1} Resizing & 112/0\%    &0.25$\times$ &73.0   &52.5    \\
    Resizing & 96/0\%     &0.18$\times$  &70.0   &49.9    \\
    \end{tabular}
    }
    \caption{\textbf{Image token reduction}.}
    \label{tab:token reduction}
\end{subtable}
\begin{subtable}[ht]{0.45\textwidth}
    \vspace{1.0em}
    \resizebox{\linewidth}{!}{
    \begin{tabular}{cccc|cc}
    Stage & Step &Step\_size &Compute & Clean & PGD-20 \\
    \shline
    \rowcolor{mygray1}\multirow{3}{*}{Pre-training} &1    &4/255   &1.0$\times$ &73.0   &52.5  \\
     & 2    &3/255   &1.5$\times$  &72.1   &52.6 \\
     &3    &3/255   &2.0$\times$  &71.8   &52.5 \\ \hline
    \multirow{3}{*}{Fine-tuning}  &1    &4/255   &1.0$\times$  &75.0   &50.6 \\
     &2    &4/255   &1.5$\times$  &73.2  &52.3  \\
     \rowcolor{mygray1} &3    &4/255   &2.0$\times$  &73.0  &52.5  \\
    \end{tabular}
    }
    \caption{\textbf{Attack strength}. }
    \label{tab:attack strength}
\end{subtable}

\begin{subtable}[ht]{0.35\textwidth}
    \vspace{1.0em}
    \resizebox{\linewidth}{!}{
    \begin{tabular}{cc|cc}
    Approach & Ratio/Size & Clean & PGD-20 \\
    \shline
    \multirow{2}{*}{w/o Tuning} & 160/0\%    &74.4   &43.2    \\
     & 112/0\%    &68.5   &39.3    \\ \hline
     & 160/0\%    &74.7   &53.9    \\
    \rowcolor{mygray1}\multirow{-2}{*}{w Tuning} &112/0\%    &73.0   &52.5    \\
    \end{tabular}
    }
    \caption{\textbf{Fine-tuning}. }
    \label{tab:full res finetune}
\end{subtable}
\caption{\textbf{Ablating design choices} with ViT-B/16 on ImageNet-1K. We report clean and PGD-20 robust accuracy (\%). If
not specified, the default setting is: \textit{112$\times$112 image size for pre-training, 224$\times$224 image size for fine-tuning; PGD-1 for pre-training, and PGD-3 for fine-tuning; 200 epochs for pre-training length, 20 epochs for fine-tuning}. Default settings are marked in \colorbox{mygray1}{gray}. In table (a) and (b), note that full-resolution fine-tuning is included. In table (b), when tuning the PGD step and step size in pre-training, we fix them to be 3 and 4/255 respectively in fine-tuning; When tuning the PGD step and step size in fine-tuning, we fix them to be 1 and 4/255 respectively in pre-training.}
\label{tab:ablation study}
\vspace{-1em}
\end{table}
\endgroup

\subsection{Design Choices}
\label{sec:design choices}
We first conduct an ablation study on the design choices of AdvXL using ViT-B/16 on ImageNet-1K, with robust accuracy under PGD-20 serving as the primary metric for adversarial robustness. 
We maintain the default baseline setting (see the caption of Tab.~\ref{tab:ablation study}). Any alterations are confined to the specific factors under examination.

\begingroup
\renewcommand{\arraystretch}{1.4} 
\begin{table*}[htbp]
\centering
\resizebox{\linewidth}{!}{
\begin{tabular}{cccccc|cc}
Case & Model        & Dataset                                     & Samples@Resolution     & Adv. Steps & \begin{tabular}[c]{@{}c@{}}Compute\\ (1e10)\end{tabular} & Clean & PGD-20 \\
\shline
Baseline &ViT-B/16     & ImageNet-1K                & 256M@112 + 38.4M@224   & 1/3        &0.5    &73.0        &52.5       \\ \hline
model scaling &\textbf{ViT-L/16}    &ImageNet-1K                                             & 256M@112 + 38.4M@224   & 1/3    &1.7    &74.8    &54.7       \\
model scaling &\textbf{ViT-H/14}    &ImageNet-1K                                             & 256M@112 + 38.4M@224   & 2/3    &5.7    &76.5    &54.9       \\
model+data scaling & \textbf{ViT-L/16}    &\multirow{1}{*}{\textbf{+ ImageNet-21K}} & 256M@112 + 38.4M@224 &1/3        &1.7     &75.8       &56.1       \\
model+data+schedule scaling & \textbf{ViT-L/16}    &\multirow{1}{*}{\textbf{+ ImageNet-21K}} & \textbf{789M}@112 + 38.4M@224 &1/3        &3.4     &77.2       &58.3       \\
model+data+schedule scaling & \textbf{ViT-H/14}    &\multirow{1}{*}{\textbf{+ ImageNet-21K}} & \textbf{789M}@84 + 38.4M@224 &2/3        &8.1     &79.0       &60.5       \\ 
\hline

model+data+schedule scaling &\textbf{ViT-L/16}    & \textbf{ + LAION-400M}                                  & \textbf{2.56B}@112 + 38.4M@224           & 1/3        &8.8      &80.5        &62.2       \\
model+data+schedule scaling &\textbf{ViT-H/14}    & \textbf{ + DataComp-1B}                                 & \textbf{5.12B}@84 + 38.4M@224           &2/3        &38.6     &83.3        &68.2   
\end{tabular}
}
\vspace{-0.5em}
\caption{\textbf{Scaling behavior of AdvXL}. For each model, we report its training set, the number of training samples it used and their resolution, its PGD number of steps (in pre-training and fine-tuning, respectively), the total training compute (in 1e10 GFLOPS), clean accuracy, and PGD-20-robust accuracy. 
``+'' on the dataset means any additional dataset used during training besides ImageNet-1K.
We scale along three aspects: model, data, and scale, and observe consistent improvement in terms of both clean accuracy and robustness. 
}
\label{tab:scaling behavior}
\end{table*}
\endgroup

\begingroup
\renewcommand{\arraystretch}{1.2}
\begin{table}[ht]
\footnotesize
\centering
    \vspace{-1em}
    \begin{tabular}{cc|cc}
    Dataset & Model & Clean & PGD-20 \\
    \shline
    \multirow{2}{*}{ImageNet-1K}   &ViT-B/16    &73.0   &52.5    \\
     & ConvNeXT-B    &73.9   &54.2    \\ \hline
    \multirow{2}{*}{+LAION-400M} &\cellcolor{mygray1} ViT-L/16    & \cellcolor{mygray1}80.5   & \cellcolor{mygray1}62.2    \\
    &ConvNeXT-L   &77.9   &58.5    \\
    \end{tabular}
    \vspace{-1em}
    \caption{\textbf{Architecture comparison} between ViT and ConvNeXT. }
    \label{tab:architecture}
    \vspace{-1em}
\end{table}
\endgroup

\begingroup
\renewcommand{\arraystretch}{1.2}
\begin{table}[ht]
\footnotesize
\centering
    \begin{tabular}{c|cc}
    Text Encoder & Clean & PGD-20 \\
    \shline
    Base &80.6  &62.2  \\
    Large &80.5  &62.2  \\
    Huge &80.6  &62.3  \\
    \end{tabular}
    \vspace{-1em}
    \caption{\textbf{CLIP text encoder size}. }
    \label{tab:clip text encoder}
    \vspace{-2em}
\end{table}
\endgroup

\paragraph{Token Reduction.} Our investigation delves into three distinct strategies for reducing image token length: 1) random masking, which randomly removes a portion of input tokens; 2) block masking, which retains a large consecutive block of the input grid; 3) resizing, which preserves most high-level semantic information. As shown in Tab.~\ref{tab:token reduction}, all three methods exhibit substantial computational speedups. Notably, image resizing demonstrates superior performance among these strategies, presumably because it suffers the least from loss of information. For instance, resizing the input image to $112\times112$ leads to a 75\% reduction in total computation, with only a minor decrease of 2.5\% in clean accuracy and 2.0\% in PGD-20 robust accuracy. \textit{We select an image size of $112\times112$ for pre-training as the default setting due to its satisfactory balance between efficiency and performance}.

\paragraph{Attack Strength.} Tab.~\ref{tab:attack strength} scrutinizes the impact of varying attack steps and step sizes during pre-training and fine-tuning. Intriguingly, we observe that the number of PGD steps for pre-training does not need to align with that for fine-tuning. For instance, adopting PGD-1 for pre-training yields nearly equivalent robustness compared to PGD-3, while reducing the computation by 100\%. This suggests that despite exposure to weaker attacks during pre-training (\eg, with PGD-1), a short-term but stronger adversarial fine-tuning (\eg, with PGD-3) is sufficient for the model to secure strong robustness against adversarial attacks. Therefore, \textit{we opt to use PGD-1 for pre-training and PGD-3 for fine-tuning in our default setting}. 

\paragraph{Fine-tuning.} Tab.~\ref{tab:full res finetune} outlines the impact of full resolution fine-tuning with stronger attacks for an extra 20 epochs on the ImageNet-1K dataset. For $112\times112$ PGD-1 pre-training, a $224\times224$ PGD-3 fine-tuning elevates clean accuracy by 4.5\% and PGD-20 robust accuracy by 13.2\%. This fine-tuning phase substantially narrows the performance gap between reduced-length pre-training and full-length training, demanding only around 60\% of the pre-training computational resources. Extending the pre-training schedule by the corresponding compute yields significantly inferior results, highlighting the distinct advantage of fine-tuning in achieving a superior performance-compute trade-off. Therefore, \textit{we consistently integrate a short-term fine-tuning stage post pre-training}.

\subsection{Scaling Behavior}
\label{sec:scaling behavior}
The acceleration outlined previously allows us to delve into the performance implications of scaling AdvXL within an affordable computational budget. In particular, we scrutinize the scaling behavior along three principal axes below, in line with the approach established by Li~\etal~\cite{li2023scaling}:

\begingroup
\setstretch{1.05}
\begin{itemize}
    \item \textit{Model scaling}. We substitute the ViT-B/16 model with ViT-L/16 or ViT-H/14, which has $\app$2$\times$ or $\app$4$\times$ number of parameters, respectively. 
    \item \textit{Data scaling}. We substitute the training set of ImageNet-1K with three much larger datasets, excessively expanding the total number of training samples up to more than $\app$1B. These datasets include ImageNet-21K~\cite{deng2009imagenet}, a superset of ImageNet-1K; LAION-400M~\cite{schuhmann2021laion}, and DataComp-1B~\cite{gadre2023datacomp}, two web-scale datasets.
    \item \textit{Schedule scaling}. To delineate the influence of large dataset size from that of extended training duration, we conduct training on ImageNet-21K with the same number of seen samples as training on ImageNet-1K.
\end{itemize}
\endgroup

By meticulously traversing these three scaling axes, we scrutinize their individual effects on AdvXL's performance. The findings are detailed in Tab.~\ref{tab:scaling behavior}, culminating in the following insights.

\paragraph{Model scaling.} The evaluation of larger model sizes reveals discernible improvements in both clean accuracy and adversarial robustness. For instance, as shown in the first and the second rows of Tab. \ref{tab:scaling behavior}, ViT-L/16 surpasses ViT-B/16 by 1.8\% clean accuracy (from 73.0\% to 74.8\%) and 1.8\% PGD-20-robust accuracy (from 52.5\% to 54.7\%) when training on ImageNet-1K.
Interestingly, ViT-H/14, despite its superior clean accuracy and tripled computational expense, demonstrates only a slightly better performance (0.2\% higher PGD-20 robustness) compared to ViT-L/16 when training on ImageNet-1K, as shown in the third row of Tab. \ref{tab:scaling behavior}. However, it notably surpasses ViT-L/16 by a substantial margin (2.2\% in PGD-20 robustness) when training on the larger ImageNet-21K dataset (as shown in the fifth and sixth rows of Tab.~\ref{tab:scaling behavior}). This observation suggests that larger models necessitate a larger training set to fully leverage their potential. This finding aligns with conclusions in prior studies~\cite{hoffmann2022an}, advocating for equivalent scaling of model size and the volume of training tokens.

\paragraph{Schedule scaling.} Initial experiments demonstrated that extending the training schedule for ViT-L/16 on ImageNet-1K yielded diminishing gains, possibly due to the comparatively ``limited'' scale of ImageNet-1K. However, results in Tab.~\ref{tab:scaling behavior} shows that with larger and more diverse datasets, training with additional samples yields non-trivial enhancements. Even with a 20$\times$ augmentation in the training schedule using a one-billion sample dataset, such as training a ViT-H/14 model on DataComp-1B for 5.12B samples (the last row of Tab. \ref{tab:scaling behavior}), there is not an observed saturation point.

\paragraph{Data scaling.} Our AdvXL also exhibits favorable outcomes with web-scale datasets LAION-400M and DataComp-1B. This trend could potentially pave the way for adversarially trained models to rival foundational models like CLIP~\cite{radford2021learning} and Flamingo~\cite{alayrac2022flamingo}.
 Notably, we find that data scaling itself is beneficial, even without a prolonged training schedule. 
As shown in the second and the fourth rows of Tab. \ref{tab:scaling behavior}, by substituting ImageNet-1K with ImageNet-21K to adversarially train ViT-L/16, we observe an uptick of 1.0\% in clean accuracy and a 1.4\% increase in robustness, notwithstanding identical training durations. 
When coupled with our preliminary findings suggesting diminished returns from extended schedules on ImageNet-1K, we conclude that the richness and diversity brought by data scaling stand as pivotal elements in the success of adversarial training at scale.

\begingroup
\renewcommand{\arraystretch}{1.4}
\begin{table*}[htbp]
\centering
\resizebox{\linewidth}{!}{
\begin{tabular}{cccccccc|cccc}
Model        & Dataset                                     & Samples@Resolution     & Pre-trained  & Adv. Steps & \begin{tabular}[c]{@{}c@{}}Params\\ (M)\end{tabular} & \begin{tabular}[c]{@{}c@{}}Compute\\ (1e10)\end{tabular} & Source & Clean & $l_{\infty}$ &$l_{2}$ &$l_{1}$ \\
\shline
RobArch-L    &\multirow{8}{*}{\textcolor{mygray2}{ImageNet-1K}} & 128M@224   &  &3    &104  &1.3      &\cite{peng2023robarch}  &73.5        &48.9       &39.5    &14.7  \\
ViT-B/16    & & 384M@224   &  &2    &87  &2.7      &\cite{rebuffi2023revisiting}  &76.6        &53.5       &-    &-  \\
ConvNeXT-B    & & 384M@224   &  &3    &89 &2.4      &\cite{liu2023comprehensive}  &76.0        &55.8       &44.7    &21.2  \\
Swin-B    & & 384M@224   &  &3    &88  &2.4      &\cite{liu2023comprehensive}  &76.2        &56.2       &47.9    &23.9  \\
ConvNeXt-B+ConvStem    & & 320M@224   &\checkmark  &3    &89 &2.0         &\cite{singh2023revisiting}  &75.2        &56.3       &49.4    &23.6  \\
ConvNeXt-L+ConvStem    &  & 128M@224 & \checkmark   &3        &198 &1.8         &\cite{singh2023revisiting}  &77.0        &57.7       &47.0    &22.2  \\
ConvNeXt-L+ConvStem    &  & 128M@224(320 eval) & \checkmark   &3        &198  &1.8         &\cite{singh2023revisiting}  &78.2        &59.4      &56.2    &33.8  \\
ConvNeXt-L    &  & 384M@224 &   &3        &198 &5.3         &\cite{liu2023comprehensive}       & 78.0 &58.5       &-    &-  \\
Swin-L    &  & 384M@224 &   &3        &197  &5.3         &\cite{liu2023comprehensive}       & 78.9 &59.6       &-    &-  \\
\hline
\rowcolor{mygray1} ViT-H/14   & + DataComp-1B                                 & 5.12B@84 + 38.4M@224 + 6.4M@336                       & & 2/3        &304 &39.6     &ours   &83.9        &69.8        &69.8    &46.0 \\
\rowcolor{mygray1} ViT-g/14    & + DataComp-1B                                 & 5.12B@84 + 38.4M@224 + 6.4M@336                       & & 2/3        &1013 &63.4  &ours   &83.9        &71.0        &70.4    &46.7
\end{tabular}
}
\caption{
\textbf{Comparison to SOTA $l_{\infty}$-robust models on ImageNet}. 
For each model we report the training set it used, the number and resolution of training samples it used, if it uses pre-trained weights or not, the number of PGD steps in AT (in pre-training and fine-tuning, respectively), the number of parameters of each model, the total training compute (in 1e$^{10}$ GFLOPS), its source, its clean accuracy and 
$l_{\infty}$, $l_{2}$, $l_{1}$-robust accuracy with $\epsilon_{\infty}$ = 4/255, $\epsilon_{2}$ = 2, $\epsilon_{1}$ = 75(AutoAttack). Note that for the model initialized with pre-trained weight, the pre-training compute is not included. For unavailable metrics of those publicly unavailable models, we use  ``-'' to fill in the blank. ``+'' on the dataset means any additional dataset used during training besides ImageNet-1K.
Our AdvXL successfully secures new state-of-the-art records on all three robustness metrics thanks to its unprecedented model and data scale.
}
\vspace{-1em}
\label{tab:sota comparison}
\end{table*}
\endgroup

\subsection{Architecture Choice}
\label{sec:architecture}
We have also ablated alternative architectures such as ConvNeXT~\cite{liu2022convnet} and Swin-Transformer~\cite{liu2021swin}, two leading backbones on RobustBench ImageNet leaderboard~\cite{liu2023comprehensive,singh2023revisiting}. 
However, our attempts to train a Swin-Transformer with reduced-size inputs posed challenges
as the feature size of the last stage may even be smaller than the window size. For example, when employing common configurations like a patch size 4$\times$4 and a window size 7$\times$7, using a 112$\times$112 input would lead to a final stage feature size of 3$\times$3. 
This mismatch hindered effective training without architectural modifications, and thus, we primarily focus on comparing ViT and ConvNeXT.

To ensure fair evaluation, we maintain consistency with the same two-stage training recipe detailed in Sec.~\ref{sec:scaling behavior} during the performance comparison. The results, presented in Tab.~\ref{tab:architecture}, demonstrate that ConvNeXT does outperform ViT on a relatively small scale.
However, this advantage diminishes as the scale increases, leading us to keep ViT as the default backbone for comparisons against other state-of-the-art models.

Also, we could adopt a larger pre-trained CLIP text encoder in contrastive learning, as it is frozen and introduces little computational overhead. Tab.~\ref{tab:clip text encoder} shows the result of training ViT-L/16 on LAION-400M with various CLIPA text encoders. As can be seen, the performance is robust to a wide range of CLIP text encoder choices. Thus, we simply use the same-scale text encoder to the image encoder (\ie a ViT-L image encoder with a Large text encoder).

\subsection{Comparison with SOTA Models}
\label{sec:sota comparison}

The comparison presented in Tab.~\ref{tab:sota comparison} evaluates our models against prior works, focusing on $l_{\infty}$ robustness at $\epsilon_{\infty}$ = 4/255. Following \cite{singh2023revisiting}, we include $l_2$ robustness at $\epsilon_{2}$ = 2 and $l_{1}$ robustness at $\epsilon_{1}$ = 75. Models listed exhibit over 80M parameters and are sorted based on their $l_{\infty}$ robustness under AutoAttack.

AdvXL emerges as the top performer owing to its unprecedented scale in adversarial training. Our highly efficient two-stage training paradigm facilitates this scalability without incurring excessive computational expenses. For instance, our largest ViT-g/14 model trained on the DataComp-1B dataset achieves outstanding results with a computing budget of merely about $12\times$ that of the previous best results from \cite{liu2023comprehensive}. Despite this relatively modest computational investment, our model outperforms them by an impressive 11.4\% in terms of $l_{\infty}$-robust accuracy under AutoAttack.
We would like to stress that training with full resolution and strong attacks on 5.12B samples, without our efficiency design, would incur  $\app20\times$ the computational cost of our approach (equating to $\app250\times$ the compute of the previous best results), rendering such an endeavor computationally infeasible.

Even more noteworthy is the exceptional generalizability showcased by our AdvXL ViT-g/14  models trained on the web-scale DataComp-1B dataset, securing $l_{2}$-robust accuracy of 70.4\% and $l_{1}$-robust accuracy of 46.7\%. 
These figures represent an absolute improvement of about 13\% over the best previous results. This observation indicates that scaling model, data, and schedule collectively not only significantly enhances robustness against known attacks but also fortifies the model against unseen attacks during training. Our findings on scaling adversarial training illuminate the path towards the evolution of next-generation robust visual models, potentially propelling the field of adversarial training into the era of foundation models.

\section{Discussion and Conclusion}
\label{sec:conclusion}

Adversarial training has traditionally been confined to small networks and datasets, predominantly ResNet-50 and CIFAR-10. Until recently, there have been few attempts to train adversarially robust models on the medium-size ImageNet-1K dataset. In this work, we break new ground by scaling adversarial training to web-scale datasets containing over 1B samples. Our AdvXL approach comprises two core components: 1) a coarse-to-fine, weak-to-strong, two-stage training paradigm to mitigate the computational cost of scaling up; 2) the utilization of a pre-trained CLIP text encoder enabling training on web-scale datasets. Through scaling along model, data, and schedule dimensions, we successfully establish a new state-of-the-art record of $l_{\infty}$-robust accuracy under AutoAttack, surpassing the previous best by a margin of $\app$10\%. Additionally, training on those gigantic datasets demonstrates increased generalizability against unseen attacks during training, aligning with observations from various foundation models~\cite{radford2018improving,radford2019language,brown2020language,alayrac2022flamingo,radford2021learning}. We envision our work as a stepping stone for adversarial training to enter the era of foundation models, inspiring further large-scale adversarial training endeavors.

\paragraph{Broad impact.} Our method delivers over 5$\times$ speedup, significantly reducing wall-clock time for training models with hundreds of millions or even billions of parameters on billion-scale datasets (\eg, on the order of thousands of TPU/GPU-days). AdvXL not only facilitates rapid prototyping and accelerated research cycles but also contributes to substantial energy and carbon emissions savings, a critical consideration in large-scale model training.

\section*{Acknowledge}
This work is partially supported by a gift from Open Philanthropy. We thank Center for AI Safety, TPU Research Cloud (TRC) program, and Google Cloud Research Credits program for supporting our computing needs.

{
    \small
    \bibliographystyle{ieeenat_fullname}
    \bibliography{main}
}

\end{document}